\pdfoutput=1

\documentclass[11pt]{article}

\usepackage[final]{acl}
\usepackage{hyperref}
\usepackage{times}
\usepackage{graphicx}

\usepackage{enumitem}
\usepackage{fancyvrb} 

\usepackage{tcolorbox}
\usepackage{lipsum}
\tcbuselibrary{skins,breakable}
\usetikzlibrary{shadings,shadows}
\usepackage{tcolorbox}
\usepackage{graphicx}
\usepackage{lipsum}
\usepackage{amsmath}
\usepackage[utf8]{inputenc}
\usepackage[T1]{fontenc}
\usepackage{graphicx}
\tcbuselibrary{skins,breakable}
\usetikzlibrary{shadings,shadows}
\newenvironment{myexampleblock}[1]{%
    \tcolorbox[beamer,%
    noparskip,
     colback=white,        
    colframe=black,       
    coltitle=white,         
    colbacktitle=darkblue,     
    title=#1,             
    title=#1]}%
    {\endtcolorbox}
\usepackage{latexsym}

\usepackage[T1]{fontenc}

\usepackage[utf8]{inputenc}

\usepackage{microtype}

\usepackage{inconsolata}

\usepackage{graphicx}
\usepackage{lipsum}  
\usepackage{subcaption}

\title{\texttt{KODIS}: A Multicultural Dispute Resolution Dialogue Corpus}

\author{James Hale$^1$, Sushrita Rakshit$^2$, Kushal Chawla$^3$, Jeanne M. Brett$^4$, Jonathan Gratch$^1$ \\ $^1$University of Southern California,  $^2$University of Michigan, \\ $^3$Capital One, $^4$Northwestern University \\ \texttt{jahale@usc.edu, gratch@ict.usc.edu} }

\begin{document}
\maketitle
\begin{abstract}
We present \texttt{KODIS}, a dyadic dispute resolution corpus containing thousands of dialogues from over 75 countries. Motivated by a theoretical model of culture and conflict, participants engage in a typical customer service dispute designed by experts to evoke strong emotions and conflict. The corpus contains a rich set of dispositional, process, and outcome measures. The initial analysis supports theories of how anger expressions lead to escalatory spirals and highlights cultural differences in emotional expression.
 We make this corpus and data collection framework available to the community\footnote{Please fill out this \href{https://docs.google.com/forms/d/e/1FAIpQLSfM1yJM0qUQQA8GQlp-k_vQscZ65Qe5NqxsxTnP8xtZ5c3keA/viewform?usp=header}{Google Form} to access the data-set.}.
\end{abstract}


\section{Introduction}

Conflicts ubiquitously arise between individuals, organizations, nations, and cultures. Conflicts can help individuals recognize and appreciate differences and learn essential social skills. Too often, conflicts escalate to verbal, legal, or physical violence~\cite{brett1998breaking, halperin2008group}. Individual conflicts can damage relationships and incur costly legal fees. National conflicts cost the global economy USD \$19 trillion in 2023~\cite{GPI24}. ``Culture wars'' within and between nations give rise to different conceptions of reality that can perpetuate generational conflict~\cite{MARSELLA2005651}.  

Interest grows in using natural language processing (NLP) methods to understand how conflicts arise and are resolved through conversation~\cite{chawla2023social,gelfandLLM24,davani2023hate}. Conflict dialogues are task-oriented non-collaborative conversations: task-oriented as parties aim to achieve specific goals (e.g., extract concessions), and goal achievement can be explicitly measured;  non-collaborative as goals are misaligned, though not necessarily zero-sum (parties could discover win-win compromises via conversation).

This paper introduces a large (4,061 participants) and novel corpus designed to offer multicultural insights into how conflicts escalate or resolve through conversation. It is novel in that we examine \textit{dispute resolution} rather than \textit{deal-making}. Deal-making is a major recent focus of NLP research~\cite{lewis-etal-2017-deal,he-etal-2018-decoupling,cheng-etal-2019-evaluating,chawla2021casino,kwon2024llmseffectivenegotiatorssystematic}, though this literature has favored the term negotiation over deal-making. This creates confusion as both deal-making and dispute-resolution involve negotiation (parties converse to influence each other and extract concessions, often over multiple issues), but disputes involve unique social processes~\cite{brett2014negotiating} and have received less attention within the fields of AI and NLP.  By prioritizing deal-making over dispute resolution, the NLP community risks overlooking key processes that shape conflict, which this corpus seeks to address.

Deal-making is forward-looking as parties focus on opportunities for gain and try to establish a new relationship. As the relationship is not yet established, parties have greater opportunities to explore alternatives. When parties fail to reach an agreement, they can seek other partners -- e.g., if unsatisfied with one car dealer, one can always negotiate with another. In contrast, disputes are backward-looking, typically involving an existing relationship that has gone badly. As parties are already linked, success depends on managing the costs of ending the relationship rather than opportunities moving forward~\cite{brett2014negotiating}. As a result, disputes evoke much stronger emotions, and positions are more entrenched than deal-making. This distinction is crucial as it shapes the consequence of influence attempts. For example, whereas expressions of anger promote compromise in deal-making~\cite{van2004interpersonal}, they evoke escalation in disputes~\cite{pruitt2007conflict, adam2015context,adam2018everything}. As disputants often become entrenched in their positions, rather than seeking compromise, they seek to overpower their opponent through appeals to justice (``you violated my rights!'') or by threatening harm (``I will sue you!''), leading to a spiral of further escalation, including threats of physical violence~\cite{brett1998breaking, halperin2008group,pruitt2007conflict}. Thus, the costs of disputes can greatly exceed the original perceived injury, engulfing not only the disputants but other related parties and even society at large.  

The corpus is also novel in using theories of conflict to guide data collection. We collected a diverse sample of participants from over 75 countries, and participants were matched within and across countries. We measure cultural and individual differences that have previously been shown to shape negotiated outcomes. For culture, we focus on differences between Dignity, Face, and Honor cultures~\cite{leung2011within,yao2017measurement}. This theoretical framework distinguishes cultures by the
degree to which people’s social identity is independent versus interdependent and thus shapes the importance given to norms of reciprocity and honesty. Dignity cultures (typically Western society) might respond to a norm violation with a shrug or even a smile. Honor cultures (typically the Middle East or South America) might respond with hot anger (especially if the violation involves family). Face cultures (typically East Asia) might react by shutting down all emotional expressions~\cite{aslani2016dignity}. Additionally, we measure individual differences such as risk-propensity~\cite{Meertens2008Measuring} and the specific goals parties bring to a dispute. Finally, we assess several theoretical mechanisms surrounding the dispute, including process variables (e.g., what tactics did parties use, what emotions were expressed, and did parties understand their partner's interests?) and outcome variables, including objective and subjective measures concerning the outcome of the dispute.

Finally, this corpus' novelty partially stems from including a mix of human-human and human-AI disputes. Though we focus on human disputes, when participants could not match with a partner promptly, they matched with a large language model (GPT-4), assuming their partner's role. Post-conflict measures include beliefs about whether their partner was human or AI and attitudes towards using AI technology for such applications. Analysis of this data could yield insight into the current limitations of GPT-4; differences between human and AI dialogs; and how these differences shape dispute processes, outcomes, and perceptions.  

\begin{figure*}
    \centering
    \includegraphics[width=.8\linewidth]{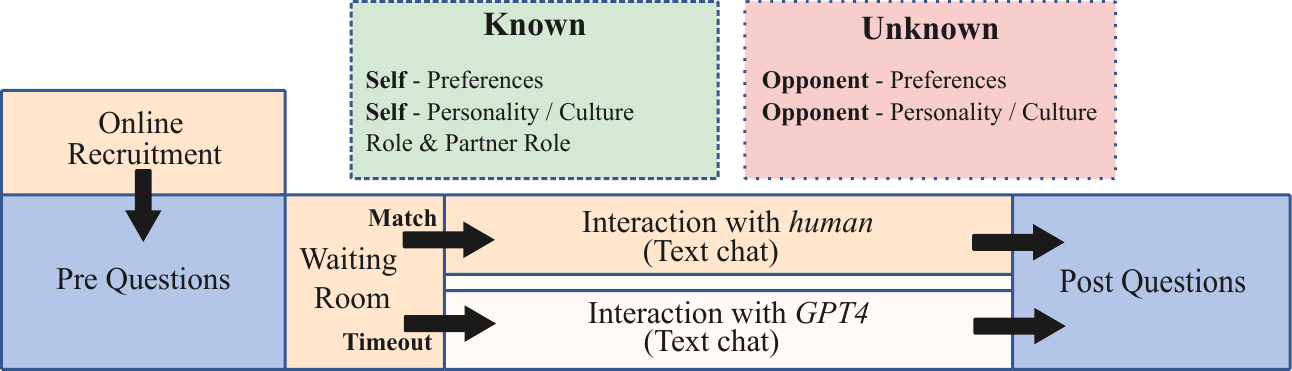}
    \caption{Participants did pre / post-dispute questionnaires and interacted with their counterpart. We first try to match the participant with a human, though we match them with GPT4 if unmatched after seven minutes.}
    \label{fig:lioness}
\end{figure*}

We envision a wide range of theoretical and applied uses for the corpus. For social scientists, it offers a test-bed to study critical social processes such as function of emotional expressions~\cite{friedman2004positive}, the role of perspective-taking~\cite{klimecki2019role,galinsky2011using}, the dynamics of escalation and deescalation~\cite{pruitt2007conflict}, and the impact of culture~\cite{tinsley2004culture,gelfand2001cultural}. From a general artificial intelligence perspective, it provides a means for evaluating AI's social competence~\cite{gratch2015negotiation,kwon2024llmseffectivenegotiatorssystematic,NutToM24} and uncovering AI's tendency to propagate cultural stereotypes~\cite{havaldar2023multilingual}. From an application perspective, it supports the development of AI agents to help teach dispute-resolution skills~\cite{gelfandLLM24,murawski2024negotiage}, to monitor and intervene in human disputes~\cite{cho-etal-2024-language}, or to replace customer service agents with AI~\cite{ebers2022automating}. 

Researchers must be mindful of ethical pitfalls when pursuing these ends. A tool like GPT can yield important insights, yet its uncritical use can adversely affect knowledge production and understanding~\cite{GPTology,messeri2024artificial}, and our GPT-informed conclusions in Section~\ref{sec:emotion} should be generalized with care. Concerning applications, AI developed to resolve conflicts could reinforce structural inequalities between cultures~\cite{lin2022artificial} or be repurposed to create conflict. As a result, we limit corpus access to non-commercial uses. 

This work contributes the following:
\begin{itemize}
    \item We first describe the creation and nature of the \texttt{KODIS} corpus, which we make available to the NLP community. 
    \item We then summarize a recent study using the corpus to illuminate how emotional expressions shape disputes as they unfold within and between cultures.
\end{itemize}


\section{\texttt{KODIS} Corpus}
We introduce \textit{KObe DISpute corpus} (\texttt{KODIS}), a corpus of dyadic disputes. Our data collection was inspired by the CaSiNo framework of \citet{chawla2021casino}, which allows pairs of human participants to match online and engage in a deal-making exercise via text chat. Prior NLP-based analysis of CaSiNo dialogues found that emotional expressions predict participants' satisfaction with their negotiated agreement~\cite{chawla2023towards}, that information expressed in the dialogue revealed participant's private goals for the negotiation~\cite{chawla2022opponent}, and that agents could be trained via reinforcement learning to negotiate effectively against people~\cite{chawla-etal-2023-selfish}. We adapt this framework to dispute resolution and extend it to allow human-agent disputes. 
\begin{figure*}
\centering
    \includegraphics[width=.9\linewidth]{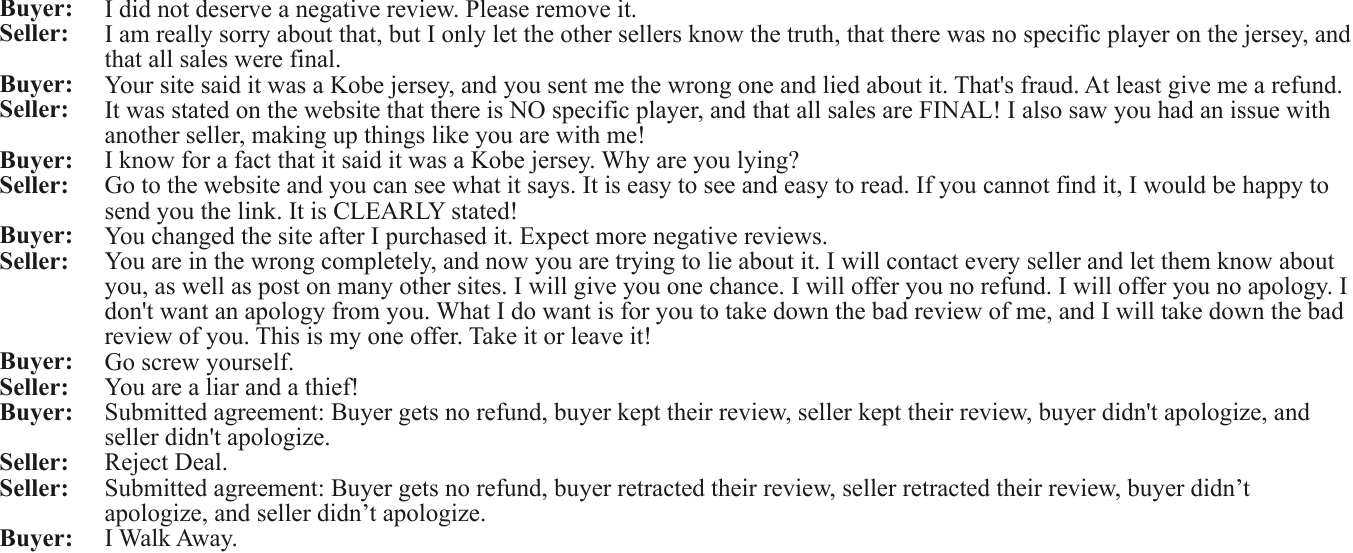}
    \caption{This illustrates an example of a contentious dialogue from the \texttt{KODIS} corpus.}
    \label{fig:dialog_ex}
\end{figure*}

\subsection{Corpus Collection Framework}
Figure~\ref{fig:lioness} illustrates the \texttt{KODIS} data collection framework, which allows dyadic text-based interaction between two people or between a person and an AI partner. Participants are recruited through an online service such as Prolific. They first enter a survey implemented in  Qualtrics (qualtrics.com). This administers the consent form, pre-task measures, and describes the dispute scenario\footnote{Note some dispute information is common knowledge between but some is private to each party.}. Participants next perform the dispute via an application created using Lioness Labs, a  framework for multi-participant behavioral experiments~\cite{giamattei2020lioness}. 

They first enter a virtual waiting room, which displays a timer counting down from eight minutes and a message asking them to wait for their partner.  If another participant with an opposing role joins in the first seven minutes, the two meet in a chat interface and work on the dispute; otherwise, at one minute left in the lobby, the participant moves on and converses with an AI counterpart.  After completing the task, participants return to Qualtrics, where they complete post-task measures. Finally, they are routed to the online recruitment service, where they enter a completion code and receive compensation. 
When performing the task, we leave the partner's nature (e.g., human or AI) ambiguous but imply to be human. At the end of the study, participants are debriefed about the partner's true nature.

\subsection{Application Architecture}
We use \citet{giamattei2020lioness}'s Lioness Lab to implement our design as it facilitates matching online participants into dyads. Other dialogue research has used Lioness~\cite{chawla2021casino}.  Figure~\ref{fig:interface} (appendix) shows the user interface  -- on the left is a chat box; 
 a menu on the right lets participants walk away or finalize their agreement.

\subsection{Dispute Resolution Task}
Dispute resolution research advances through a mixture of field studies and scenario studies that afford measurement and manipulation of theoretically relevant variables. Here, we adopt the latter approach by designing a task to validate a theoretical model of multi-cultural negotiation proposed by \citet{aslani2016dignity}. This model argues that culture shapes negotiation tactics (e.g., competitive vs. cooperative dialog) and proposes how this relates to creating win-win solutions.

Participants engage in a role-playing exercise simulating a bilateral multi-issue buyer-seller dispute --- other dyadic corpora also use role-play settings \cite{chawla2021casino, yamaguchi2021dialogue, lewis-etal-2017-deal, he-etal-2018-decoupling}. We crafted the exercise in collaboration with an expert in dispute resolution (one of the authors) and designed it to evoke strong emotions and entrenched positions while adhering to the ethical guidelines for human experimental research. The dispute centers on a buyer who purchased a basketball jersey for their sick nephew and claims they received the wrong item. The seller disputes this claim, arguing the correct item was sent and no refunds are allowed. Each side is told the other has posted negative reviews attacking their counterpart's reputation.  Each side receives different versions of events to encourage a dispute over facts.

We frame the task as multi-issue, where participants can discuss and potentially reach a compromise. Participants are told they can discuss the refund, drop their review, request their opponent drop their review, and discuss who, if anyone, should apologize. In the appendix, Figure~\ref{fig:BuyerInt} illustrates the context presented to a participant slated to act as a buyer, and Figure~\ref{fig:sellerInt} illustrates the context presented to the seller. Figure~\ref{fig:dialog_ex} demonstrates an example dialogue where participants display hostility to one another.

Participants are incentivized to take the dispute seriously by offering a substantial bonus based on performance (up to \$3). Participants receive a base pay (\$3.50) for attempting the task, which can be nearly doubled if they achieve all their objectives. Many role-playing scenarios use ``assigned preferences,'' meaning participants are given a payoff matrix that defines their goals in the negotiation (e.g., they might receive the most bonus if they achieve a refund). To allow cultural variability, we use ``elicited preferences,'' meaning that participants are provided a fixed number of points to allocate across the four issues: e.g., buyers might assign 70\% of their points to receiving a refund but 30\% to receiving an apology. They receive a bonus based on their stated preferences and the actual resolution of the dispute, receiving a fixed bonus (\$0.70) if the dispute ends in an impasse. Specifically, we calculate the proportion of total possible points a participant scored (see Equation~\ref{fun:score}) and grant that proportion of the bonus.   


Participants use a chat interface to communicate. A menu interface reminds them of the issues under discussion and is used to unambiguously specify the final agreement (see Figure~\ref{fig:interface} in the appendix). After exchanging at least eight messages, they use this menu to send a final offer. Their partner can accept, reject, or counter this offer. After eight messages, Participants can use this interface to ``walk away'' from the dispute -- we refer to these as impasses. 
Buyers go first (unless one participant is an AI, in which case the AI always goes first). Participants alternate sending messages; the participant's interface is inactive while awaiting a new message.


\subsection{Measures}
We measure several theoretical constructs claimed to shape negotiation processes and outcomes, focusing on those used in research on multicultural deal-making ~\cite{aslani2016dignity}. This research suggests that cultural and individual variables (e.g., Dignity, Face, and Honor) shape negotiation tactics (e.g., tendency to express emotion, willingness to exchange information). Tactics shape perspective-taking (e.g., less sharing means less understanding of your opponent's goals), shaping the likelihood of reaching an agreement and the quality of the resulting agreement. Agreement quality can be measured in objective terms, but subjective feelings about the agreement and partner are more predictive of subsequent behavior, such as willingness to follow through on agreements and maintain a future relationship with the partner~\cite{curhan2006people,brown2012utility}

\subsubsection{Pre-Dispute Measures}
\textbf{Screening:}
Participants are screened using techniques shown to improve data quality~\cite{QualtricsCommitment}. They first respond to a \textit{commitment request} asking if they could commit to providing thoughtful responses. They next respond to a series of attention checks, including multiple choice questions (how many legs does a cat have?) and open-ended responses (describe the flavor of a tomato). Participants who fail these checks are immediately excluded without compensation.

\textbf{Questionnaires:}
Participants complete a demographic survey that includes gender, education level, and country where they spent most of their life (and the number of those years). 
They next complete an 18-item Dignity, Face, and Honor scale measuring cultural differences in the perception of identity threats \cite{leung2011within}. Dignity cultures view individuals as having intrinsic self-worth, making identity relatively impervious to attack. Honor cultures view self-worth as something that must be claimed and defended from external threats. Face cultures also see self-worth as conferred by others but see retaliation as further eroding self-worth. Participants are asked to consider their culture and answer questions on the strength of attitudes on a 7-point Likert scale. Examples include ``People must always be ready to defend their honor.'' and ``People should be very humble to maintain good relationships.'' 
Finally, Participants complete a short measure of risk propensity \cite{Meertens2008Measuring}. 

\textbf{Preference Elicitation:}
After reading the scenario, participants self-report their goals for the upcoming dispute by allocating 100 points over four issues to be discussed (see Figure \ref{fig:pref} -- \textit{refund}, \textit{other drops negative review}, \textit{you drop negative review}, and \textit{receive apology}). Points indicate the utility of fully or partially achieving this goal and directly map onto the participant's monetary bonus. For example, assigning more points to an apology than to a refund indicates that the participant would experience greater reward from receiving an apology than a refund. Participants are also asked to write a one-sentence justification for the importance assigned to each issue.  For example, one participant wrote, ``He's a crook and will defraud others'' as a justification for wanting to keep his bad review of the other party, suggesting they assign importance to reputational concerns.




\textbf{Integrative Potential:}
 The preferences of the Buyer and Seller determine the structure of the dispute. Depending on each party's interests, there may be an opportunity to ``grow the pie'' by finding mutually beneficial solutions. For example, if the Buyer only cares about a refund and the Seller only wants the Buyer to drop their negative review, both sides can maximize their bonus. However, just because a win-win solution exists doesn't mean the parties can find it.  The concept of ``integrative potential'' measures the potential for joint gains (which may or may not be realized). 
 We operationalize integrative potential using the preferences elicitation from each side. Given two vectors of preferences $\Vec{X}_{buyer}$ and $\Vec{X}_{seller}$: 
 \begin{align}
     \text{IP} = 1 - \dfrac{\Vec{X}_{buyer} \cdot \Vec{X}_{seller}}{\|\Vec{X}_{buyer} \| \| \Vec{X}_{seller}\|}
 \end{align}
Notably, above, we invert the cosine similarity of the two vectors --- those with dissimilar values have greater potential for joint gains. 

\subsubsection{Post-Dispute Measures}
\textbf{Perspective-Taking:}
Participants are asked a series of questions to assess if they accurately understand their partner's goals in the dispute. Participants are first asked about their preferences as an attention check. They rank each issue's importance (most, middle, or least). This is contrasted with preferences they provided during the preference elicitation phase to check if they recall their initial preferences. They are asked to do the same for their partner. The distance between these estimates and their partner's actual preferences can serve as a measure of perspective-taking accuracy.

\textbf{Tactics: }
Participants are asked a 10-item questionnaire \cite{aslani2016dignity} about the tactics they and their partner used during the dispute. On a five-point Likert scale, participants answer if they agree or disagree with a series of questions about the dispute process. E.g., ``I expressed frustration,'' ``The OTHER PARTY expressed frustration,'' and ``I shared my preferences with the other party.''
Thus, we also capture first and second-person annotations --- i.e., participants annotated the emotions and negotiating tactics of themselves and their counterpart. 

\textbf{Subjective Value of the Outcome:}
Subjective perceptions of the outcome of a dispute are a better predictor of future negotiation decisions than the actual economic result~\cite{brown2012utility}. We capture these impressions with an 8-item version of the Subjective Value Inventory (SVI)~\cite{curhan2006people}.  This measures four dimensions of subjective value, including feelings about the \textit{instrumental outcome} (e.g., ``Did I get a good deal?''), the \textit{process} (e.g., ``Was the process fair?''), feelings about the \textit{self} (``Did I lose face?''), and  feelings about \textit{relationship} with the partner (``Would I work with my partner again?'').

\textbf{Objective Individual Outcome:}
We calculate the objective value of each side's outcome from the agreed-upon outcome using the points provided in the preference-elicitation phase. 
These points are also used to determine the bonus.
Specifically, the objective value of the outcome is derived using the following linear additive utility function: 
\begin{align}\label{fun:score}
    U_a = \sum_{i\in I} w_{i,a} * \ell_{i,a}
\end{align}
Where $U_a$ denotes the points participant $a$ earns; $I$ represents the set of all issues; $w_{i,a}$ holds the value participant $a$ gave to issue $i\in I$; and $\ell_{i,a} \in [0,1]$ holds the level agreed upon for $i$ in the dispute (where higher is more favorable to $a$) --- typically binary, though the \textit{refund} issue uses .5 for a partial refund. Notable, this formula only applies in disputes ending in agreement --- those resulting in an impasse yield a fixed amount to each side.

\textbf{Objective Joint Outcome:}
 We can measure the collective benefit two parties achieve by summing the individual outcomes ($U_a+U_b$). If parties are effective negotiators, the joint outcome should be positively correlated with the integrative potential for the task, but if parties misunderstand each other (e.g., come from different cultures), they may fail to realize this potential.




\subsection{Participants and Corpus Characteristics}
We outline the demographic and dispositional composition of the corpus. 
\begin{figure}
    \centering
    \includegraphics[width=.8\linewidth]{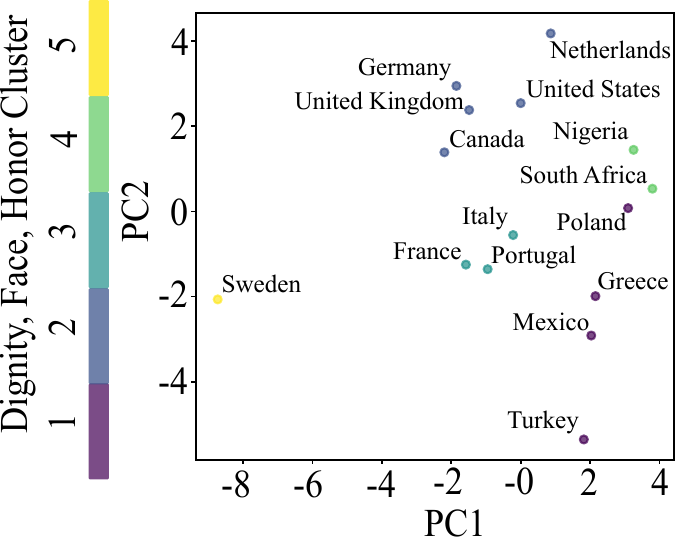}
    \caption{K-means clusters of countries ($N\geq 10$).}
    \label{fig:kmeans}
\end{figure}
\subsubsection{Demographics}
We collected responses from the crowd-sourcing platform Prolific from November 2023 to June 2024. 
We recruited participants from countries worldwide (see  Table~\ref{tab:particpants}). Further, our dataset comprises of 50\% Female, 49\% Male, and 1\% other. 
\begin{table}[h]
\small
\centering
\begin{tabular}{|rccc|}

\hline
\multicolumn{1}{|l}{\textbf{}}                          & \multicolumn{2}{c}{\textbf{Dyad}}          & \multicolumn{1}{l|}{}             \\
\multicolumn{1}{|c}{\textbf{Country}}                   & \textbf{Within} & \textbf{Mixed}           & \multicolumn{1}{l|}{\textbf{LLM}} \\ \hline \hline
\multicolumn{1}{|r|}{United States}          & 826             & \multicolumn{1}{c|}{116} & \multicolumn{1}{c|}{498}                                  \\
\multicolumn{1}{|r|}{United Kingdom}                    & 148             & \multicolumn{1}{c|}{90}  &         \multicolumn{1}{c|}{41}                          \\
\multicolumn{1}{|r|}{Canada}                            & 135             & \multicolumn{1}{c|}{75}  &    \multicolumn{1}{c|}{15}                               \\
\multicolumn{1}{|r|}{Mexico}                            & 96              & \multicolumn{1}{c|}{62}  &       \multicolumn{1}{c|}{95}                           \\
\multicolumn{1}{|r|}{South Africa}                      & 84              & \multicolumn{1}{c|}{43}  &         \multicolumn{1}{c|}{51}                          \\
\multicolumn{1}{|r|}{Portugal}                          & 9               & \multicolumn{1}{c|}{39}  &           \multicolumn{1}{c|}{5}                        \\
\multicolumn{1}{|r|}{Poland}                            & 6               & \multicolumn{1}{c|}{27}  &        \multicolumn{1}{c|}{10}                           \\
\multicolumn{1}{|r|}{Italy}                             & 3               & \multicolumn{1}{c|}{13}  &      \multicolumn{1}{c|}{4}                             \\
\multicolumn{1}{|r|}{Netherlands}                       & 3               & \multicolumn{1}{c|}{22}  &        \multicolumn{1}{c|}{0}                           \\
\multicolumn{1}{|r|}{France}                            & 1               & \multicolumn{1}{c|}{20}  &           \multicolumn{1}{c|}{2}                        \\
\multicolumn{1}{|r|}{Germany}                           & 1               & \multicolumn{1}{c|}{14}  &         \multicolumn{1}{c|}{4}                          \\
\multicolumn{1}{|r|}{Nigeria}                           & 1               & \multicolumn{1}{c|}{17}  &       \multicolumn{1}{c|}{9}                            \\
\multicolumn{1}{|r|}{Sweden}                            & 1               & \multicolumn{1}{c|}{11}  &   \multicolumn{1}{c|}{0}                                \\
\multicolumn{1}{|r|}{ \textless{}10 appearances} & 1               & \multicolumn{1}{c|}{133} &      \multicolumn{1}{c|}{15}                             \\ \hline
\end{tabular}
\caption{Counts of dyads and AI disputes collected. }
\label{tab:particpants}
\end{table}

\subsubsection{Dignity, Face, Honor, \& Risk}
We analyze the correlations between the various dispositional measures we captured --- specifically, Dignity, Face, Honor, and Risk. We see Dignity significantly positively correlates with Face ($r=.17$, $p<.001$), and Risk-seeking ($r=.13$, $p<.001$); Face significantly positively correlates with Honor ($r=.07$, $p<.01$) and Risk ($r=.05$, $p<.05$); and Honor significantly correlates with Risk ($r=.22$, $p<.001$). The recruited pool contains respondents of varying Dignity ($M=31.22$, $SD=4.03$), Face ($M=25.95$, $SD=5.15$), Honor ($M=24.33$, $SD=7.04$), and Risk-seeking ($M=4.03$, $SD=1.67$) propensities.

\subsubsection{Culture}
We attempt to delineate culture using the previously described Dignity, Face, Honor scale, using K-means to cluster similar countries.  
We only consider countries with $N \geq 10$ appearances in the dyads. Using the elbow method, we run K-means using five clusters --- Figure~\ref{fig:kmeans} depicts the resultant clusters. For easy visualization, the $x$ and $y$ axis of this figure use PCA decomposition to show the otherwise three-dimensional data; Dignity has positive loadings with PC1 (0.63) and PC2 (0.77); Face has a negative (-0.18) loading with PC1 and a positive (0.05) one with PC2; and Honor has a positive (0.75) loading with PC1 and a negative (-0.63) one with PC2.
We see the first cluster comprised of Poland, Greece, Mexico, and Turkey; the second has the Netherlands, Germany, the United Kingdom, the United States, and Canada; the third has Italy, France, and Portugal; the fourth has Nigeria, and South Africa; while the fifth has Sweden.  



\section{Using Emotion to Analyze Disputes}~\label{sec:emotion}
We summarize an initial use of the corpus to provide insight into how emotions and culture shape dispute processes and outcomes. This highlights an example use case and demonstrates the corpus has ``face validity'' in that we replicate findings from the dispute resolution literature. 
 Specifically, we use an LLM to objectively quantify the emotions expressed in the dialogue and analyze how this relates to objective (impasse vs. agreement) and subjective (subjective value inventory) outcomes.

Disputes often evoke strong emotions like anger, and expressed anger has different social consequences in disputes than in deal-making.  In deal-making, expressions of anger often signal that one side has reached their limit and the other must make concessions~\cite{van2004interpersonal}. Thus, expressions of anger can lead to concession-making. In contrast, anger often provokes escalation in disputes~\cite{pruitt2007conflict,adam2015context,adam2018everything}. Culture also plays a role in the consequences of expressions. In dignity cultures, expressions of anger are often viewed as acceptable expressions of self-interest, whereas anger can provoke retaliation in cultures where self-worth is conferred by others~\cite{adam2010cultural}.

We use the \texttt{KODIS}  to address several theoretical claims about the role of emotion in disputes: do expressions of anger provoke escalation and impasses in disputes (as previously claimed), can negotiation satisfaction be predicted by emotional expression alone, and how does culture shape these findings? 
\begin{figure}[!htb]
    \centering
        \centering
        \includegraphics[width=.9\linewidth]{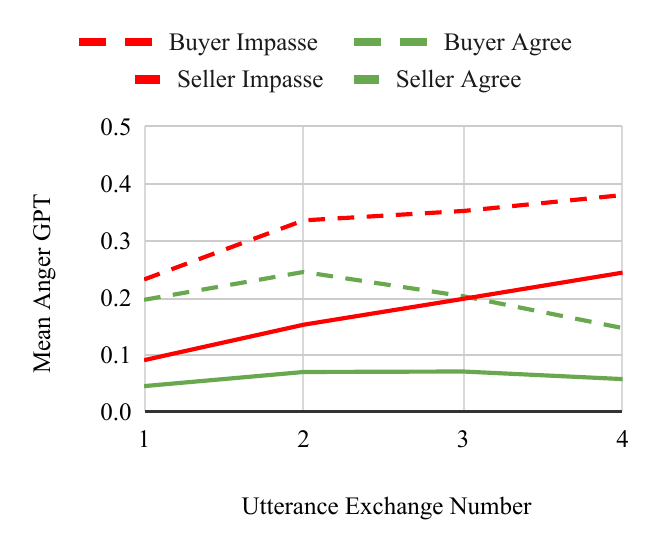}
        \caption{Anger by role, outcome and dialogue turn.}
        \label{fig:angertime}
\end{figure}

\begin{figure*}
    \includegraphics[width=\linewidth]{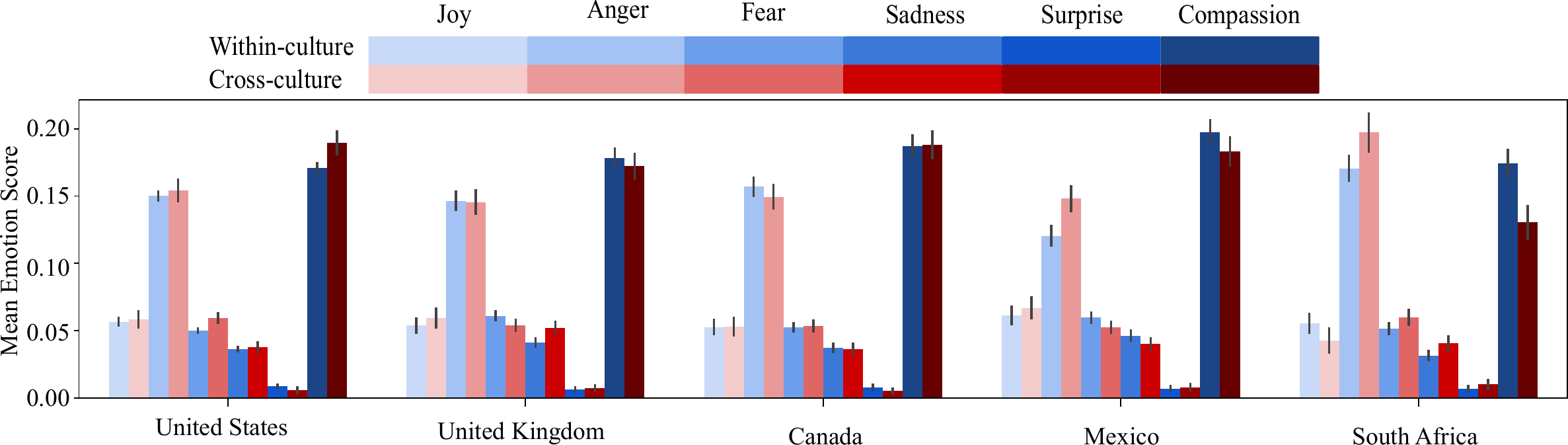}
    \caption{Average GPT emotion scores for the five most common countries broken by within or cross-culture.}
    \label{fig:cont}
\end{figure*}
\subsection{Approach}
Prior research on the CaSiNo corpus used NLP techniques to examine the impact of emotion in deal-making, and we closely follow that approach~\cite{chawla2023towards}, allowing qualitative comparisons with a similar deal-making corpus. Following research demonstrating that GPT-4 currently yields the most accurate inferences on negotiation dialoges~\cite{kwon2024llmseffectivenegotiatorssystematic}, we limit our reported analysis to GPT-4o. 

We prompt GPT4o (run on 06/28/2024) \cite{achiam2023gpt} to annotate each dialogue turn. Following detailed experiments on different prompting methods and emotion labels, GPT annotates the intensity ([0..1]) of six emotions expressed in the turn (anger, compassion, sadness, joy, fear, and surprise). Anger and compassion, in particular, were chosen based on their role in prior negotiation research~\cite{van2004interpersonal,Allred97compassion,TingToomey14-compassion}. Each utterance is presented along with the preceding dialogue context, and a small amount of in-context learning guides responses. Further, GPT reports results in a machine-readable JSON format. See the appendix for justification of these choices and evidence that GPT substantially improves predictive accuracy compared to earlier methods. For brevity, we only examine human-human dyads.

\subsection{Escalation}

The negotiation literature suggests that disputes are more likely to end with non-agreement than deal-making due to increased anger and entrenched positions.  By comparing \texttt{KODIS} disputes with the CaSiNo deal-making dialogues, we replicate this finding: about 18\% of the \texttt{KODIS} dialogues ended with an impasse (one side walked away) whereas only ~3.5\% of the CaSiNo dialogues ended with an impasse. This is remarkable as participants forfeit a cash bonus if they fail to achieve an agreement, even though this was merely a simulated dispute.

To examine escalatory dynamics, we divide dialogues based on whether they ended in an agreement or impasse. We then examine expressed emotion by role over time (see Figure~\ref{fig:angertime}). Dialogues that ended in an impasse are in red; those that ended in an agreement are in green. The results show evidence of escalation. Buyers generally enter the dialogue with greater anger. Dialogues that end with an impasse show evidence of escalation:  sellers reciprocate anger, leading to even greater anger by the buyer. In contrast, sellers avoid reciprocating anger for dialogues ending in agreement, and buyers subsequently express less anger. This supports work on conflict spirals~\cite{pruitt2007conflict}.

\subsection{Predicting Subjective Outcome}
\begin{figure}
    \centering
    \includegraphics[width=.9\linewidth]{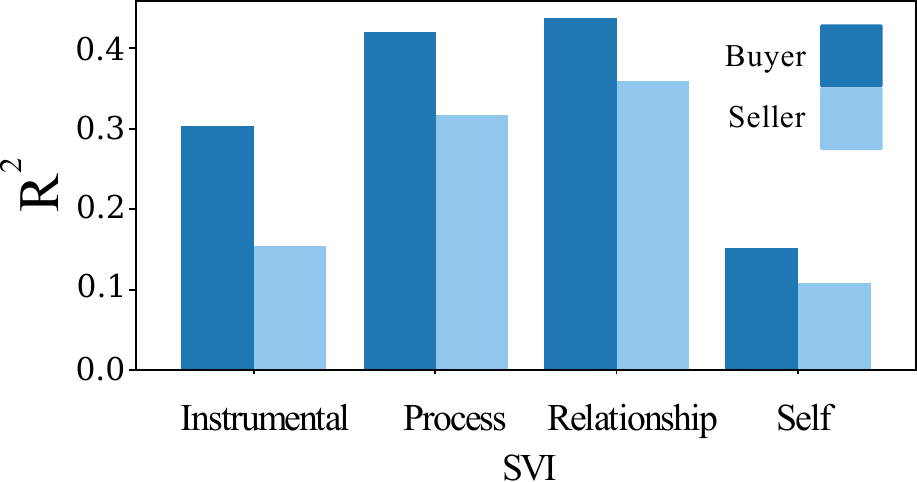}
    \caption{$R^2$ using GPT emotion to predict outcome.}
    \label{fig:SVIR2}
\end{figure}

We next examine if emotions expressed during the dialogue can predict participants' subjective feelings about the result. 
We used multiple linear regression on a random subset of 406 dialogs to predict the four facets of subjective value from expressed emotion. These include feelings about the outcome, self (did I lose face?), process (was it fair?), and relationship with the partner. We construct one model over the entire dataset to assess the relationship between emotion and subjective outcomes. 
Overall, expressions are surprisingly accurate at predicting subjective feelings. Examining $R^2$ of the regression models, emotional expressions predict almost 50\% of the variance in feelings about the process and the partner. This is particularly remarkable as these models ignore the content of the dialog. Regarding the coefficients, all emotions play a significant role in the model prediction (see Figure~\ref{fig:SVIR2}).

We additionally examine how culture shapes these predictions, which could lead to systematic bias, by training a model on Western culture and testing its predictive accuracy on other cultures. Given imbalances in our dataset, we use the nation as a proxy for culture and focus our analysis on the most prevalent countries in our sample (US, UK, Canada, Mexico, and South Africa). 
Specifically, we trained a model on a random 50/50 split of US v. US dialogues and tested on pure and mixed-nation dialogues from the other countries and unseen US data --- we ran this 1,000 times on different splits of the US-US dyads and Table~\ref{tab:cross-culture-result} displays resultant differences (averaging across all runs and SVI measures). We see the US within-culture regression explain the variance of the subjective outcome better for the countries similar and worse for dissimilar ones --- as uncovered during the cultural K-means.
Lastly, Figure~\ref{fig:cont} illustrates the differences in emotions for the five countries we consider when disputing against the same or a different country.




\begin{table}[]
\small
\centering
\begin{tabular}{|rcc|}
\hline
\multicolumn{1}{|l}{\textbf{}}                           & \multicolumn{2}{c|}{\textbf{$R^2$}}                            \\
\multicolumn{1}{|c}{\textbf{Country}}                    & \textbf{Within-culture}      & \textbf{Cross-culture}       \\ \hline \hline
\multicolumn{1}{|r|}{{\color[HTML]{000000} US}}          & {\color[HTML]{000000} 0.236} & {\color[HTML]{000000} 0.227} \\
\multicolumn{1}{|r|}{{\color[HTML]{000000} UK}}          & {\color[HTML]{000000} 0.188} & {\color[HTML]{000000} 0.215} \\
\multicolumn{1}{|r|}{{\color[HTML]{000000} Canada}}      & {\color[HTML]{000000} 0.211} & {\color[HTML]{000000} 0.214} \\
\multicolumn{1}{|r|}{{\color[HTML]{000000} Mexico}}      & {\color[HTML]{000000} 0.137} & {\color[HTML]{000000} 0.170} \\
\multicolumn{1}{|r|}{{\color[HTML]{000000} SouthAfrica}} & {\color[HTML]{000000} 0.047} & {\color[HTML]{000000} 0.157} \\ \hline
\end{tabular}
\caption{US regression on mixed and pure dyads.}
\label{tab:cross-culture-result}
\end{table}










\section{Conclusion \& Future Work}
We collected a corpus of human disputes and showed some promise using NLP methods to illuminate processes. Current analyses predict cultural tendencies from the dialogues and develop models to assist human disputants (e.g., can an algorithm recognize an escalatory spiral and suggest an intervention to help parties reach an agreement?). The corpus continues to expand, focusing on additional languages and refining the theoretical measures.

\section{Limitations}
We design the corpus to address existing limitations in the literature on NLP and negotiation by emphasizing the distinction between dispute resolution and deal-making and providing a substantial corpus of disputes. However, it is important to emphasize that the data is drawn from a single artificial scenario. The dispute focused on a standard consumer economic dispute, and participants were asked to role-play. Thus, care must be taken in generalizing these findings to real-world interactions. 

This analysis of cultural differences contains confounds that must be unpacked. We predict dispute outcomes from expressed emotion and show that performance degrades when models trained on US participants are applied to non-US participants. Yet it is unclear if this is due to bias in emotion recognition (e.g., does GPT-4o over-estimate anger in South African dialogues) or if emotion functions differently in different cultures.  Evidence from other research suggests both factors are probably in play. Thus, the analysis should be augmented by human annotations from those target cultures.

Though we collect information on demographics, personality, and culture, all information comes from self-reports, which we cannot verify. Participants act as part of a paid service, which may shape their responses in ways that do not match real-world interactions.
Participants also retained anonymity during the interaction, which can shape their responses. In real interactions, parties have an existing relationship, and there can be real-world consequences. These factors can strongly shape the expression and function of emotions.

We use commercial pre-trained models to recognize emotional expressions in our dialogues. Still, we do not have independent human annotations of what emotions will likely be perceived in the text. So, while we provide some evidence of external validity (expressed emotion impacts outcomes in theoretically predicted ways), subsequent research must verify these machine-generated emotions correspond to the intended mechanism. This is particularly fraught in a cross-cultural setting as existing work shows that large pre-trained models introduce bias in interpretations~\cite{havaldar2023multilingual}. 

\section{Ethical Considerations}
\textbf{Data Collection}
Our study was approved by USC's Institutional Review Board. The participant received informed consent, describing the purpose of the study, data policies, and noting that they could withdraw at any time. Participants could lose their compensation by withdrawing which can be seen as coercive, but aligned with current experimental norms and designed to ensure that experimental protocols match real-world decision-making~\cite{EcoExperiments}. The compensation was set to provide a fair wage and to conform to the guidelines of the online collection service. 
No identifiable information was collected during the collection. Potentially identifiable information such as IP addresses, worker IDs, and location information is removed before releasing the data. Any mention of demographics or personality of participants is based on self-identified information in our pre-survey and standard procedures of collecting personality metrics.

\textbf{Potential Risks}
Our work supports using NLP methods to provide insight into psychological processes. However, there are reasonable concerns that NLP can undermine the diversity of scientific research (by over-reliance on a small number of tools), create the illusion of objectivity, and reinforce cultural stereotypes~\cite{messeri2024artificial}. This is particularly the case for research on emotion.  Recent findings in affective science emphasize that emotions are perhaps best seen as cultural constructs labeled and interpreted differently across cultures. Yet many labeling schemes used in emotion databases rely on Western representational taxonomies. This is true of the labels we adopted in the evaluation experiment. This can serve to reinforce Western biases on the interpretation of the data. As noted by \citet{GPTology}, using LLMs ``as an off-the-shelf `one-size-fits-all' method in psychological text analysis—can lead to a proliferation of low-quality research, especially if the convenience of using LLMs such as ChatGPT leads researchers to rely too heavily on them.'' Augmenting our findings with diverse models and human judgments remains imperative. 



\section*{Acknowledgments}
    This work is supported by the U.S. Government including the Air Force Office of Scientific Research (grant FA9550-23-1-0320), and the National Science Foundation (grant 2150187). The views and conclusions contained in this document are those of the authors and should not be interpreted as representing the official policies, either expressed or implied, of the Army Research Office or the U.S. Government. The U.S. Government is authorized to reproduce and distribute reprints for Government purposes notwithstanding any copyright notation herein. We thank Jessie Hoegen, Kelly Tsai, and Jasmine Zhu for their contributions. 
\bibliography{custom}

\appendix

\label{sec:appendix}

\section{GPT4 Emotion Labelling}

\subsection {Construct Validity}\label{sec:construct}
We check whether our GPT emotion labels, compared with T5 \cite{raffel2020exploring}, accurately capture expressed emotion. We do not have ground truth, however we do have self-reported frustration. Thus, we assess how well each predicted emotion correlates with this self-report.
We see GPT-4o outperform T5 in the magnitude and direction of correlation, as seen in Table~\ref{frustration-emotion}.

\begin{table}[h!]
\small
\centering
\begin{tabular}{|l|c|c|}
\hline
\multicolumn{1}{|c|}{\textbf{Emotion}} & \textbf{\begin{tabular}[c]{@{}c@{}}T5-Twitter\\ Frustration\end{tabular}} & \textbf{\begin{tabular}[c]{@{}c@{}}GPT-4o\\ Frustration\end{tabular}} \\ \hline
Anger                                  & 0.509                                                                     & 0.553                                                                 \\
Fear                                   & -0.010                                                                    & 0.371                                                                 \\
Sadness                                & 0.120                                                                     & 0.178                                                                 \\
Surprise                               & -0.001                                                                    & -0.021                                                                \\
Compassion                             & -                                                                         & -0.201                                                                \\
Love                                   & -0.390                                                                    & -                                                                     \\
Joy                                    & -0.435                                                                    & -0.349                                                                \\ \hline
\end{tabular}
\caption{Correlation coefficients of self-reported frustration with emotion scores between T5-Twitter and GPT-4o.}
\label{frustration-emotion}
\end{table}

\subsection{Predicting Subjective Outcomes}

\begin{figure}[tbh!]
    \centering
    \includegraphics[width=.8\linewidth]{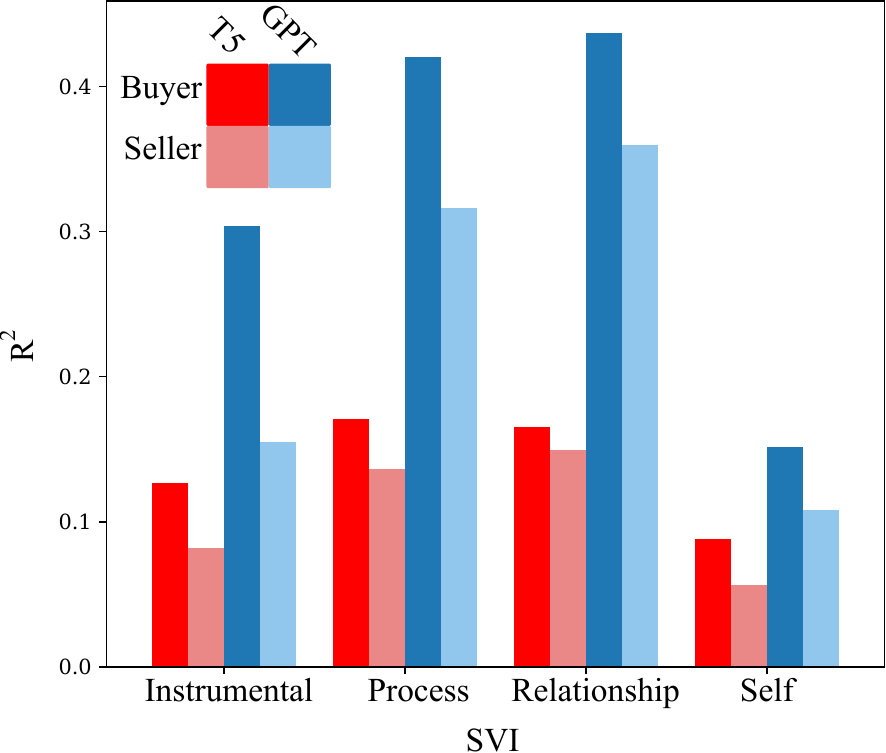}
    \caption{$R^2$ predicting the four subjective value inventory (SVI) sub-scales using GPT or T5 emotion labels}
    \label{fig:SVI_MLR}.
\end{figure}

For a random sample of 406 dialogues, we regressed the SVI scales on the emotion scores. We analyzed how well the model fits the data ($R^2$) with GPT4 scores compared to T5.
Figure~\ref{fig:SVI_MLR} depicts the varying $R^2$ values across the different configurations. The biggest leap in fit ($R^2$) comes from using GPT-4o rather than T5, as GPT explains almost half the variance in several measures. 

\subsection{GPT Prompt}
Figure~\ref{fig:prompt_text} outlines the prompt used for GPT4o in the emotion labeling task. 

\begin{figure*}[tbh!]
\centering
\begin{myexampleblock}{LLM Emotion Classifier Prompt}
\small{
You are a good emotion classification
tool. Your task is to classify the emotion of the last speaker based on the contextual dialogue.\\

Your output should be a JSON object with an `emotion' field, categorizing the dialogue with a score for each: joy, anger, fear,
sadness, surprise, compassion, or neutral. These scores should sum to one. If an utterance is neutral, then neutral must be one with everything other label set to zero.\\

Here are a few examples of proper annotations:
\texttt{
\begin{itemize}[label={}]
    \item \{"statement": "Hi ! I ’d like to return my jersey.", "emotion": \{"joy": "0", "anger": "0", " fear ": "0", "sadness": "0", "surprise": "0", "compassion": "0", "neutral": "1"\}\},
    \item \{"statement": "Please understand this was for my dear nephew he loves Kobe. I understand we had a misunderstanding, last thing I want is
to hurt your business. Let’s resolve this together", "emotion": \{"joy": "0", "anger": "0", "fear": "0.4", "sadness": "0", "surprise":
"0", "compassion": "0.6", "neutral": "0"\}\},
    \item \{"statement": "Thank you!", "emotion": \{"joy": "1", "anger": "0", "fear": "0", "sadness": "0", "surprise": "0", "compassion ": "0", "
neutral ": "0"\}\},
    \item \{" statement ": "I will report you to authorities for doing this .", "
emotion ": \{"joy": "0", "anger": "1", "fear ": "0", "sadness ": "0", "surprise": "0", "compassion ": "0", "neutral": "0"\}\}
\end{itemize}}

}
\end{myexampleblock}
    \caption{This outlines the prompt GPT used in the emotion annotation task.}
    \label{fig:prompt_text}
\end{figure*}

\section{Task Background}
\begin{figure*}
    \includegraphics[width=\linewidth]{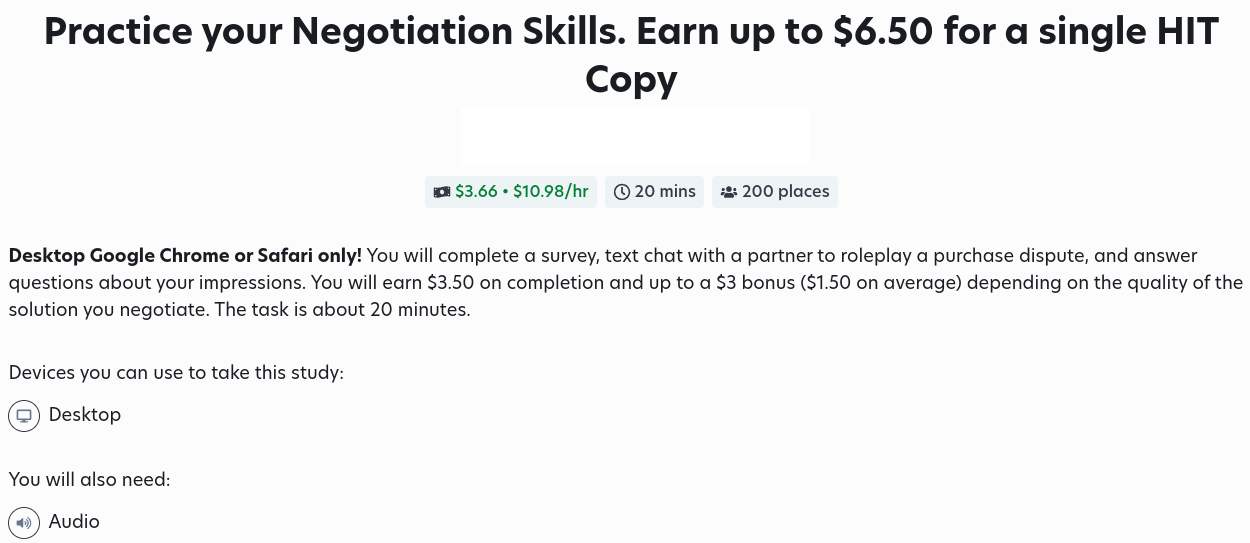}
    \caption{Depiction of the Prolific recruitment page for crowdworkers. }
    \label{fig:prolific}
\end{figure*}

\begin{figure*}[h]
    \centering
    \begin{subfigure}[b]{0.48\textwidth}
        \centering
        \includegraphics[width=\textwidth]{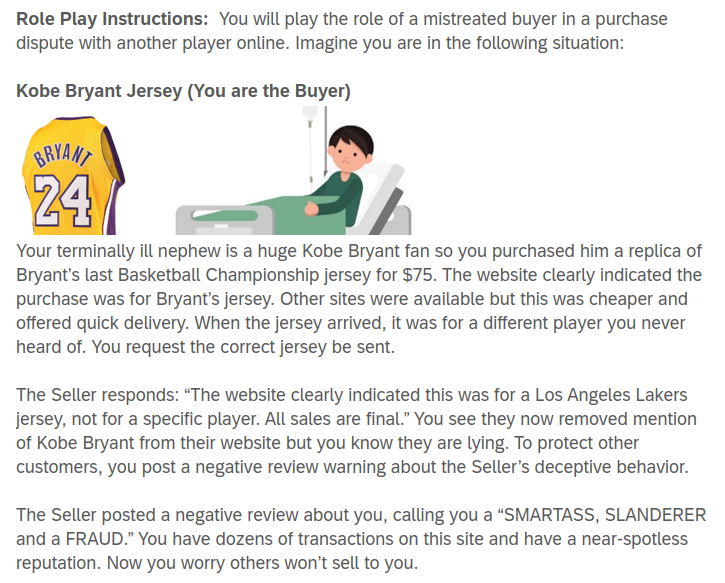}
        \caption{Buyer}
        \label{fig:BuyerInt}
    \end{subfigure}
    \hfill
    \begin{subfigure}[b]{0.48\textwidth}
        \centering
        \includegraphics[width=\textwidth]{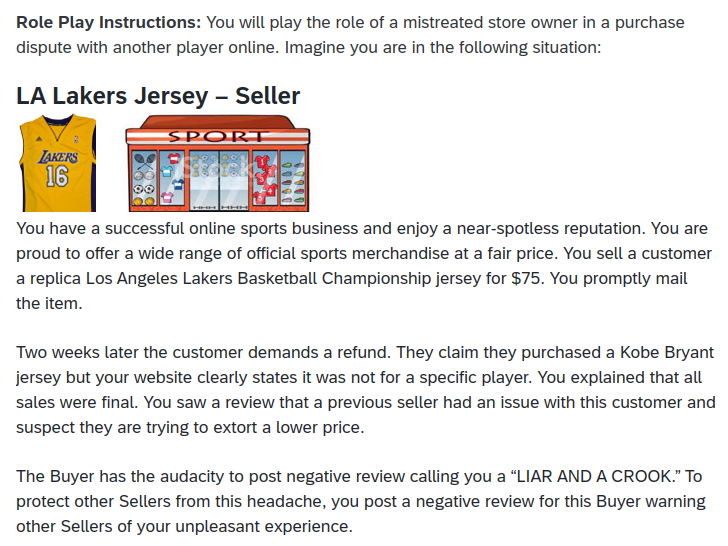}
        \caption{Seller}
        \label{fig:sellerInt}
    \end{subfigure}
    \caption{This illustrates the role-play instructions for the buyer and seller, which participants read before engaging the the dispute.}
    \label{fig:instructions}
\end{figure*}

\begin{figure*}[h]
    \centering
    \begin{subfigure}[b]{0.8\textwidth}
        \centering
        \includegraphics[width=\textwidth]{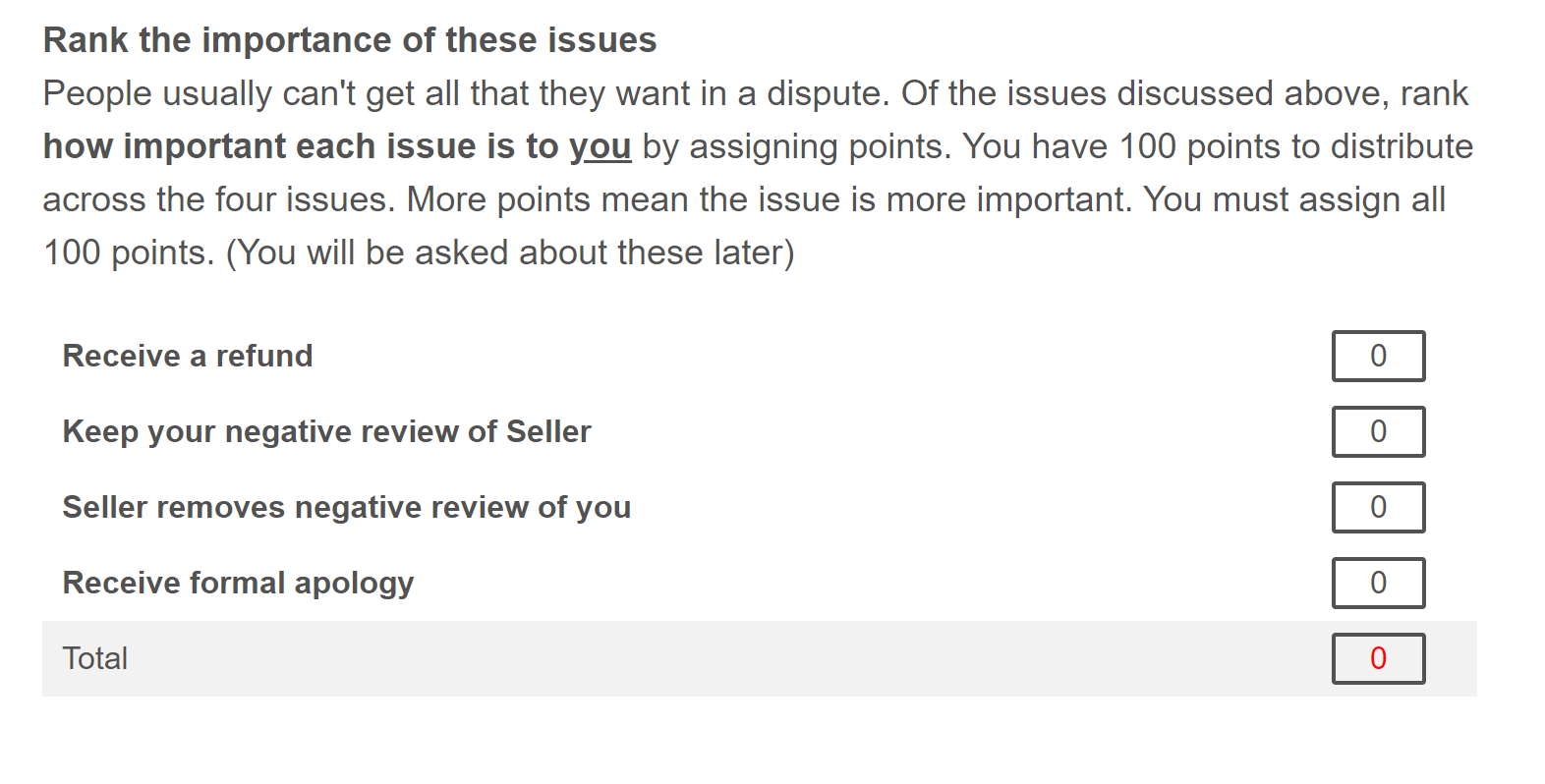}
        \caption{Participants allocate 100 points between the four issues depending on their relative importance. }
        \label{fig:pref}
    \end{subfigure}
    \hfill
    \begin{subfigure}[b]{0.8\textwidth}
        \centering
        \includegraphics[width=\textwidth]{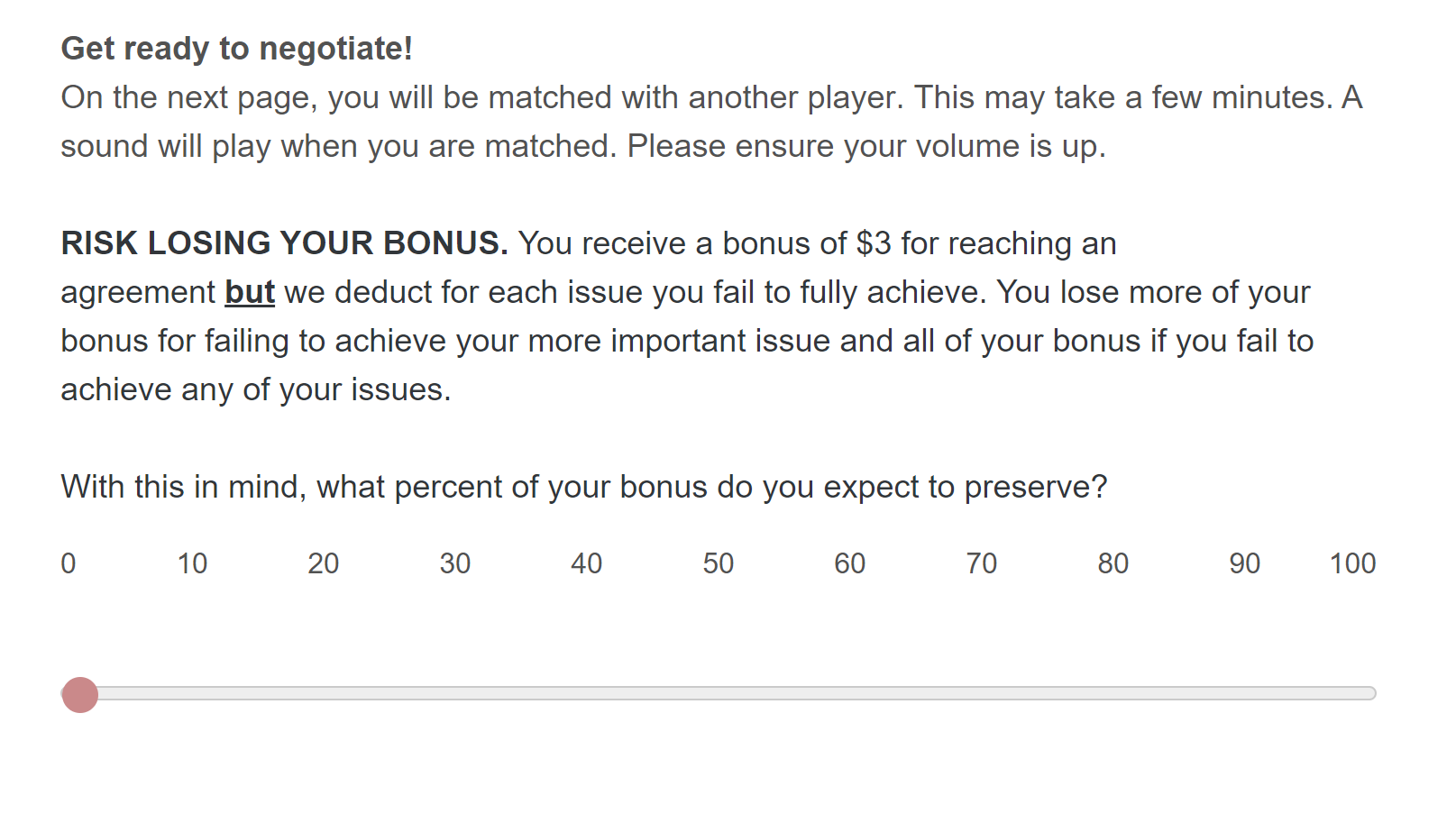}
        \caption{A slider to measure a participant's aspiration.}
        \label{fig:asp}
    \end{subfigure}
    \caption{Mechanisms for participants to input pre-dispute responses -- their preferences and aspirations.}
    \label{fig:prefasp}
\end{figure*}

This section of the appendix provides further details about the instructions for the participants, as well as the interface.  Figure~\ref{fig:prolific} depicts what the crowd worker would have seen when recruited to the task. 
Figure~\ref{fig:instructions} shows instructions the participants read before the task.
Figure~\ref{fig:prefasp} shows the interfaces through which participants would enter pre-task measures, such as their preferences or aspirations. 

\begin{figure*}[tbh!]
    \includegraphics[width=\linewidth]{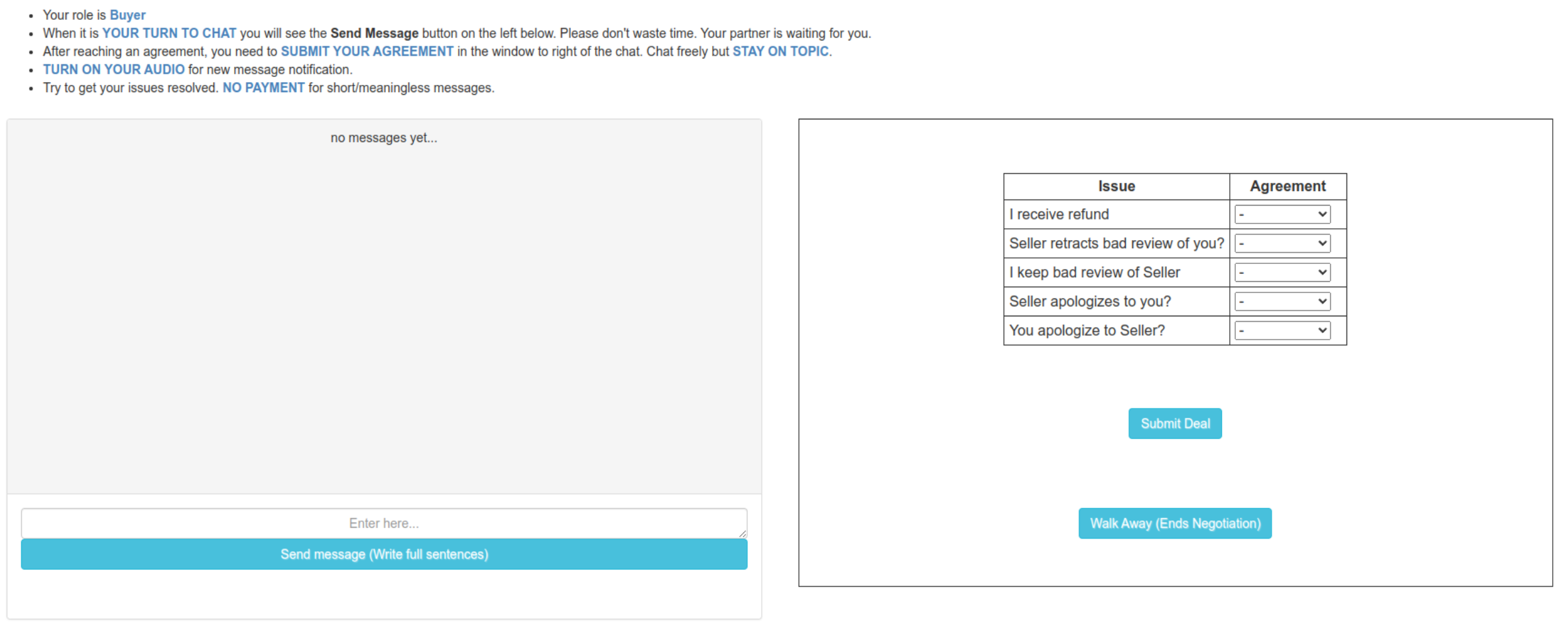}
    \caption{This depicts the interface participants used in the data collection from Lioness Labs. }
        \label{fig:interface}
\end{figure*}

\end{document}